\documentclass[10pt,twocolumn,letterpaper]{article}

\usepackage{cvpr}
\usepackage{times}
\usepackage{epsfig}
\usepackage{graphicx}
\usepackage{amsmath}
\usepackage{amssymb}
\usepackage{calrsfs}
\usepackage{subcaption}
\usepackage{algorithm}
\usepackage{algorithmic}
\usepackage{lipsum}
\usepackage[switch]{lineno}

\usepackage{setspace}

\DeclareMathAlphabet{\pazocal}{OMS}{zplm}{m}{n}
\newcommand{\Lb}{\pazocal{L}}

\DeclareMathOperator{\dplus}{+\kern -0.4em+}

\usepackage{caption,subcaption}
\usepackage{multirow}
\usepackage{bbm}
\usepackage{booktabs}
\usepackage{amssymb}
\usepackage{amsfonts}
\usepackage{dsfont}
\usepackage{mathtools}
\usepackage[inline]{enumitem}
\usepackage{amsthm}

\usepackage[pagebackref=true,breaklinks=true,letterpaper=true,colorlinks,bookmarks=false]{hyperref}

\cvprfinalcopy 

\ifcvprfinal\fi

\begin{document}

\title{Deep Multimodal Feature Encoding for Video Ordering}
\author{
    {  Vivek Sharma$^{1,3}$, Makarand Tapaswi$^{2,4}$ and Rainer Stiefelhagen$^{1}$}\\
    {\normalsize {$^{1}$Karlsruhe Institute of Technology,  $^{2}$University of Toronto, $^{3}$Massachusetts Institute of Technology, $^{4}$Inria}} \\ 
     \tt\small  \{firstname.lastname\}@kit.edu,  makarand@cs.toronto.edu, vvsharma@mit.edu
}

\maketitle

\begin{abstract}
   True understanding of videos comes from a joint analysis of all its modalities: the video frames, the audio track, and any accompanying text such as closed captions.
We present a way to learn a compact multimodal feature representation that encodes all these modalities.
Our model parameters are learned through a proxy task of inferring the temporal ordering of a set of unordered videos in a timeline.
To this end, we create a new multimodal dataset for temporal ordering that consists of approximately 30K scenes (2-6 clips per scene) based on the ``Large Scale Movie Description Challenge".
We analyze and evaluate the individual and joint modalities on three challenging tasks:
(i) inferring the temporal ordering of a set of videos; and
(ii) action recognition.
We demonstrate empirically that multimodal representations are indeed complementary, and can play a key role in improving the performance of many applications. The datasets and code are available at \href{https://github.com/vivoutlaw/tcbp}{\textit{https://github.com/vivoutlaw/tcbp}}.

\end{abstract}

\section{Introduction}

As humans, watching a movie entails looking at the video frames, listening to the audio track (both music and speech), and optionally, reading the closed captions to obtain a better understanding.
All these modalities contribute by bringing together different pieces of information.
Similarly, a joint analysis of these modalities is crucial for many practical applications, such as video retrieval, surveillance video analysis, automatic description or search within home movies and online uploaded video clips, describing video content for visually impaired, and more.
Inspired by this, multimodal feature representation learning~\cite{ngiam2011multimodal,mcb,mm1,mm2,mm3,mm4,mm5,mm6,liu2018learn} has attracted quite a lot of attention.
These multimodal sources often consist of complementary information for objects, scenes, events, dialogs, and activities of interest. 
Therefore, learning a joint representation from multiple sources is useful for robust inference.

\begin{figure}[t]
\centering
{\includegraphics[width=0.99\columnwidth]{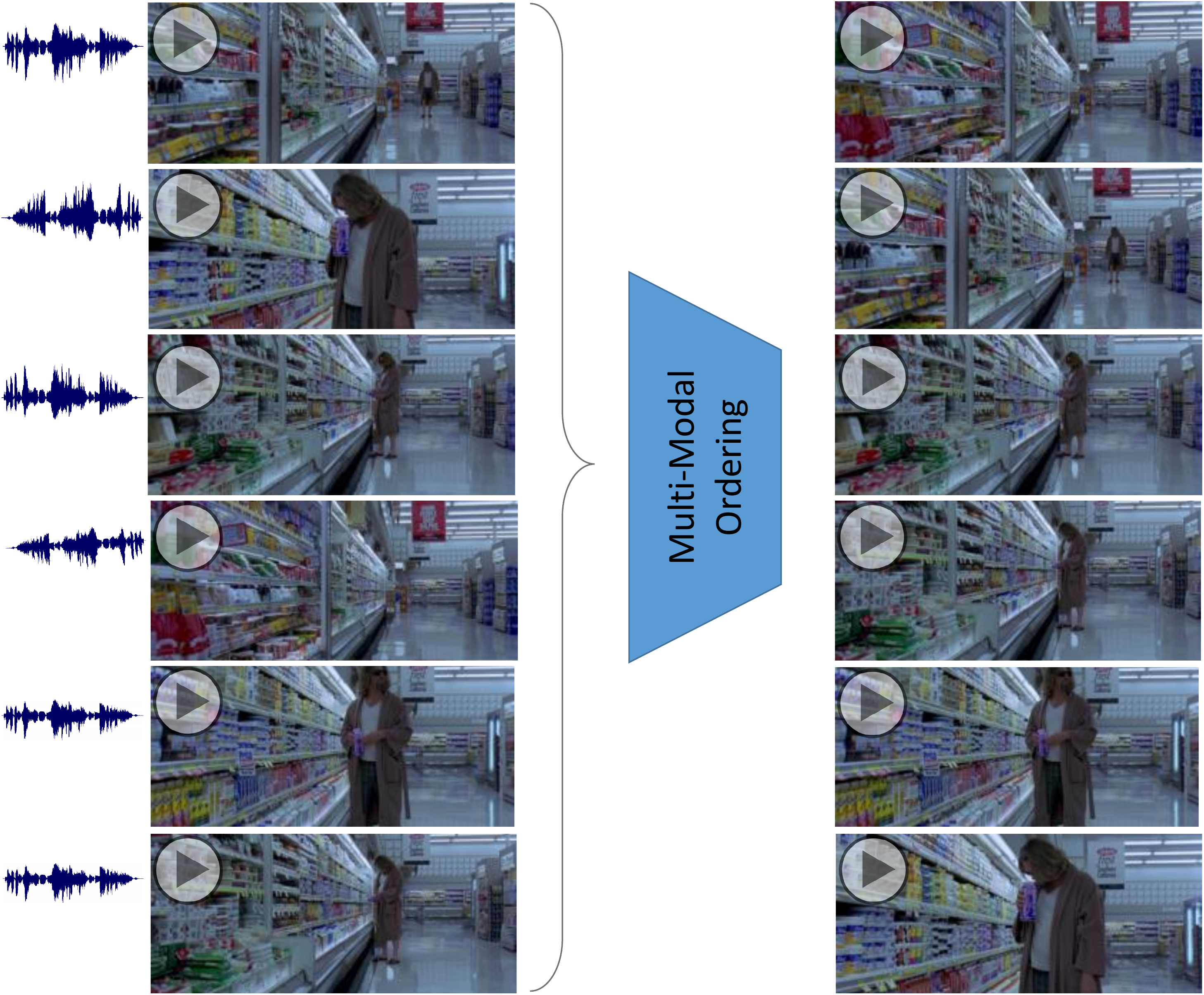} }
\vspace{-0.25cm}
\caption{Given an unordered collection of video clips, we infer their correct temporal ordering by utilizing our compact multimodal feature encoding that exploits high-level semantic concepts such as objects, scenes, and sounds in each clip.}
\label{fig:front}
\vspace{-0.6cm}
\end{figure}

Joint representations of images and text are quite popular in captioning and retrieval and are commonly achieved through a combination of convolutional neural networks (CNN)~\cite{he2016deep} and recurrent neural networks~(RNN)~\cite{kiros2015skip}.
Typical fusion strategies to bring the two domains together include element-wise product, sum, avg, max pooling or concatenation, or self-attention.
Recently, \emph{bilinear pooling}~\cite{mm6,mcb} has gained interest as it efficiently captures pairwise interactions between two representations.
Typical implementations use either the Tensor Sketch projection algorithm~\cite{tsalgo,compactbilinear} or matrix factorization~\cite{mm6}.
Beyond images and text, a few other works have investigated the combination of modalities~\cite{ngiam2011multimodal,mm2,mm3,mm4}: video and audio, or video and text.

We are interested in capturing such pairwise interactions between modalities for video clips, and propose a new deep multimodal feature encoding inspired by~\cite{oldbilinearmodels,compactbilinear,tsalgo}.
The design of the multimodal feature aims to aggregate  audio, visual, video, and text sources into a compact feature encoding.
To this end, we propose Temporal Compact Bilinear Pooling~(TCBP) as an extension of the Tensor Sketch projection algorithm~\cite{tsalgo} to incorporate a temporal dimension.
TCBP is a temporal encoding of several segments of multimodal features into a compact representation, backed by deep pre-trained models. 
Specifically, TCBP captures the interaction of each modality with one-another at all spatial locations over a temporal context (typically 2-5 seconds).


We are also motivated by self-supervised visual representation learning methods~\cite{oddoneout,shufflelearn,sharmaacmmm,sharmafg,wang2015unsupervised}, where millions of positive/negative training image pairs are generated from video constraints to learn discriminative features.
Common proxy tasks include predicting the next frame, or picking the ``odd'' one among a set of frames.
Generating such training data at the frame-level is made possible due to the stationary context and objects in the short-term temporal neighborhood.
However, we are unaware of any prior work that obtains strong supervision of constraints for obtaining clip-level training data.  
In practice, these self-supervised constraints do not scale to learn clip-level representations often due to the lack of millions of annotated videos, and also owing to limited computational resources. 
In this work we present a simple yet surprisingly powerful proxy task of temporal ordering of several clips to learn a self-supervised clip representation. 
We utilize the compact temporal multimodal feature representation for the problem of inferring the temporal ordering of video-clips. 

We highlight our key contributions below: \\
1) We propose a new self-supervised learning paradigm that produces a compact multimodal video clip representation combining audio, visual, text and video modalities. In particular, we extend the Tensor Sketch projection algorithm to incorporate the temporal dimension, and name it Temporal Compact Bilinear Pooling~(TCBP). \\
2) We introduce temporal ordering of video clips as a novel proxy task for learning clip representations. To this end, we create a new dataset of almost 30K annotated ordered scenes from movies each consisting of 2-6 clips per scene. \\
3) We present an in-depth analysis of individual and joint modalities and their effects on performance through the temporal ordering. 
We also demonstrate that TCBP can lead to performance gains in action recognition by effectively combining temporal cues.




\section{Related Work}
\label{sec:related}

\vspace{1mm}
\noindent\textbf{Self-supervised representation learning} is a line of research growing in popularity in recent years.
This paradigm includes obtaining supervision from within the structure of the data, and thus removes the need for an often costly labeling effort.
A strong cue to learn good image representations is the spatial consistency of images~\cite{doersch}.
Another approach uses a host of spatial and color transforms to train a model to recognize instances of transformed images~\cite{dosovitskiy}. 
In particular, face image representations are also fine-tuned for specific applications such as video face clustering by considering positive/negative pairs within video sub-structures~\cite{tapaswi2014icvgip} or mining labels using a distance matrix~\cite{sharmafg}.

Videos are often used to learn good image representations.
Object tracking in a video can be used to generate several instances of positive/negative pairs~\cite{wang2015unsupervised}.
At the frame level, using temporal consistency within neighboring frames~\cite{shufflelearn}; or finding an odd frame in a collection of many frames~\cite{oddoneout} are examples of methods used to automatically create positive/negative training data.
Note that many of these methods train CNN models from scratch to learn an effective representation.

In contrast to these, our work differs substantially in scope and technical approach.
We are interested in learning multimodal video clip representations (and not single images) and propose to use temporal ordering of a variable set of clips to learn our joint representation.

\vspace{1mm}
\noindent
\textbf{Time as a supervisory signal.}
The idea of using temporal continuity or ordering as a signal for supervision is quite popular.
For example, Misra~\etal~\cite{shufflelearn} shuffle a sequence of 3 frames from a video to generate positive/negative samples and train a model to predict whether a given sequence is ordered correctly.
However, the use of short 3-frame sequences only allows learning image representations, while we wish to learn multimodal clip representations.

Similarly, Wei~\etal~\cite{wei} predict whether a video flows forwards or backwards given a stack of 10 optical flow frames from a short clip.
However, this formulation prevents the method from learning how to encode object information, and requires computationally intensive optical flow maps.
Most importantly, only a binary forward/backward ordering is learned, and other shuffled permutations are not considered.
In our task of temporal ordering, we \emph{order video clips} and not frames, thus allowing to learn clip-level representations.
We order up to 6 clips each with a duration ranging from 2-5 seconds.

It is also worth noting that temporal ordering has been used for other applications such as predicting what happens next in egocentric videos~\cite{zhou}, using the steadiness of visual change in videos to penalize higher order differences~\cite{jayaraman}, or to recognize complex, long-term activities~\cite{laxton}.

Similar in spirit to our temporal ordering of video clips, arranging photos by time has also received some attention.
In particular, automating the process of creating temporally ordered photo albums from an unordered image collection~\cite{sadeghi}; sorting photo collections spanning many years~\cite{matzen,schindler}; or ordering a set of photos taken from uncalibrated cameras~\cite{basha,dicle,moses} are all related works.
In contrast, we work with an unordered collection of video clips.

\vspace{1mm}
\noindent\textbf{Audio-visual learning.}
Starting with the work by Yuhas~\etal~\cite{yuhas1989integration} which employs neural networks for audio-visual speech recognition, audio-visual learning has come a long way.
Recently, there has been interest in predicting the audio modality given a visual input~\cite{owens2016visually}, or predicting the ambient sound that an image conveys~\cite{owens2016ambient}.
There is also work on learning representations in a joint space.
For example, Aytar~\etal~\cite{soundnet} trained a student-teacher network to transfer the knowledge between the outputs of visual and audio networks.
Subsequently, these ideas were extended to learning inputs either with image+text and image+sound pairs~\cite{Aytar17}.
Audio-visual representation learning has seen applications in several tasks such as event classification~\cite{parekh2013weakly}, audio-visual localization~\cite{parekh2013weakly,arandjelovic2017objects}, biometric matching~\cite{nagrani2018seeing}, sound localization~\cite{owenseccv,arandjelovic2017objects}, person identification~\cite{nagrani2018learnable}, action recognition~\cite{owenseccv}, on/off-screen audio separation~\cite{cocktail,owenseccv}, and video captioning/description~\cite{mm1,ngiam2011multimodal,mm3}.

\section{Learning a Clip Representation}
\label{sec:app}



\begin{figure*}[h]
\centering
{\includegraphics[angle=270,width=1.99\columnwidth]{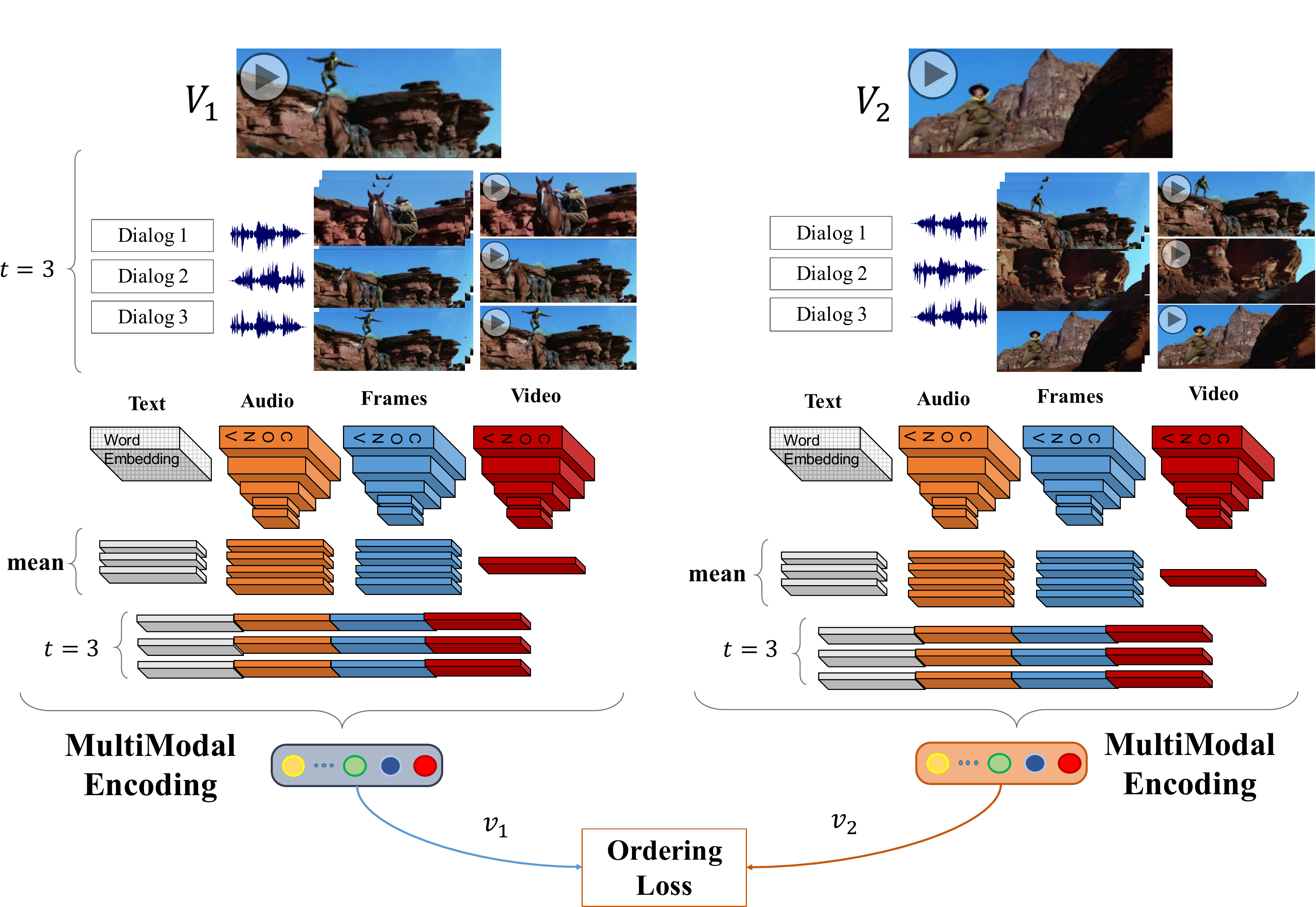}} 
\caption{\textbf{Deep Multimodal Feature Encoding.} Illustration of the multimodal feature encoding applied for task of temporal ordering problem. See Sec.~\ref{sec:app} for a detailed explanation of the feature learning scheme shown.} 
\label{fig:pipeline}
\vspace{-0.1cm}
\end{figure*}

We present our model to learn a multimodal video clip representation that encodes images, video, audio, and text jointly.
Our goal is to create a representation that is compact and efficiently encodes all modalities into a single descriptor.
We train this model in a self-supervised paradigm, with supervision from the proxy task of video clips ordering .

\subsection{Deep Multimodal Feature Encoding}
\label{subsec:mfe}
For a video clip with $N$ modalities, we compute one output feature map per modality produced by a pre-trained (convolutional) neural network.
We denote these feature maps as matrices $\{S_{1}, S_{2}, \ldots, S_{N}\}$, with size $S_{i} \in \mathbb{R}^{c_i \times t}$, $c_i$ denotes the number of channels for the $i$th modality, and $t$ is the temporal length or number of segments in which the video is divided for feature extraction.
Note that as our clips can some times be quite long (about 5 seconds) we divide them into segments of a fixed duration to compute features.


We aggregate all feature maps by concatenating them across channels, to compute $x = S_1 \dplus S_2 \dplus \ldots \dplus S_N$, where $\dplus$ represents the concatenation operator and obtain $x \in \mathbb{R}^{c \times t}$, where $c = \sum_{i=1}^{N} c_i$.
Concatenation along the channels allows us to combine a variable number of modalities. 
Additionally it ensures that information from all modalities is preserved at this stage of feature encoding.
The aggregated feature $x$ is then passed through some encoding method $E: x\rightarrow v$, to compute the multimodal encoded feature $v \in \mathbb{R}^{d}$, where $d$ is the dimensionality of the multimodal embedding space.



\subsection{Temporal Compact Bilinear Pooling}
In this work, we utilize bilinear models~(or bilinear pooling) for learning multimodal representations.
We choose to use bilinear models as they possess the ability to capture interactions between each element of the feature with one-another (much like a polynomial kernel), thus leading to a powerful feature representation.
Additionally, they have been shown to work well for related tasks of image classification~\cite{compactbilinear,bilinearmodels,chowdhury2016one}, video classification~\cite{tle} and visual-question answering~\cite{mcb}.
We will first introduce bilinear models, and then propose our extension for temporal video data.


\vspace{1mm}
\noindent\textbf{Preliminaries.}
A bilinear pooling function is an operator on the outer product of a vector $x \in \mathbb{R}^c$ to obtain $v \in \mathbb{R}^d$ such that:
\begin{equation}
v = W[x \otimes x^T] \, ,
\end{equation}
where $\otimes$ denotes the outer product, $[\cdot]$ linearizes the matrix, and $W \in \mathbb{R}^{c^2 \times d}$ are model parameters to be learned.
However, as feature dimensions are often high (e.g., $c = 1024$), the number of parameters for $W$ can easily be in the range of billions (assuming $d = 1024$ as well), which makes learning challenging.

To alleviate this problem, Gao~\etal~\cite{compactbilinear} show how the  high-dimensional second-order function can be approximated by a low dimensional projection function using the Tensor Sketch algorithm~\cite{tsalgo}.
This removes the need for computation of the expensive outer product and obtains the final vector directly.


\vspace{1mm}
\noindent\textbf{Tensor Sketch}~(TS)~\cite{tsalgo}
also called the Count Sketch algorithm, projects a vector $x \in \mathbb{R}^c$ to $v \in \mathbb{R}^d$.
Two sets of auxiliary vectors $h$ and $s$ are randomly initialized from a uniform distribution (and held fixed thereafter) to perform this projection $u_1 = \psi(x, h_1, s_1)$ and $u_2 = \psi(x, h_2, s_2)$.
The sign vector $s \in \{-1, 1\}^c$ indicates whether the element in $x$ will be added or subtracted from the final value at a random location determined by $h \in \{1,\ldots,d\}^{c}$.
In particular, for every element $x[i]$, its destination in $v$ is given by $j = h[i]$, and the value in $v$ is accumulated as $u[j] = u[j] + s[i] \cdot x[i]$.
Finally, $u_1$ and $u_2$ (corresponding to $h_1, s_1$ and $h_2, s_2$) are convolved with each other to compute $v$ by performing element-wise multiplication in the Fourier domain.


\vspace{1mm}
\noindent\textbf{Compact Bilinear Pooling} (CBP)~\cite{compactbilinear}.
When working with images, a CNN feature map $x$ often preserves the encoding of spatial locations to yield $x \in \mathbb{R}^{c \times s}$, where $c$ is number of channels (as in the above discussion) and $s$ refers to spatial locations.
To work with such features, \cite{compactbilinear} first performs TS projection on the vector at each spatial element $s$, followed by sum pooling.
In summary,
\begin{equation}
x \in \, \mathbb{R}^{c \times s} \, \xrightarrow{\text{TS projection}} \, \mathbb{R}^{d \times s} \, \xrightarrow{\text{sum pool over } s} \, v \in \mathbb{R}^{d} \, .
\end{equation}
We will refer to this process of projecting individual vectors followed by sum pooling as CBP.
For more details please refer to~\cite{compactbilinear}.

\vspace{1mm}
\noindent\textbf{Temporal Compact Bilinear Pooling (TCBP).}
We now propose an extension to the Tensor Sketch projection algorithm to incorporate a temporal dimension.
Recall, $x \in \mathbb{R}^{c \times t}$ is the aggregated feature after the concatentation, which we will encode with TCBP.
The key difference between CBP and TCBP is the process in which the TS algorithm's projection matrices $h$ and $s$ are created.
In particular, we initialize $h \in \{1, \ldots, d\}^c$ as a $c$ dimensional vector, while $s \in \{-1,1\}^{c \times t}$ as a matrix.
Having $h$ to be independent of time ensures that a feature index $i$ in $x, x[i,t]$ is always encoded to the same destination $j = h[i]$, albeit with a different sign.

Similar to the TS paragraph from above we can write $j = h[i]$ (and is independent of $t$), while, $u[j] = u[j] + \sum_t s[i,t] \cdot x[i,t]$.
Our TCBP approach is summarized in Algorithm~\ref{algo:ttsp}.

Note that the above procedure is different from applying CBP directly on a flattened vector $x \in \mathbb{R}^{ct}$, as this would require creation of $h$ which is also initialized to be $h \in \{1, \ldots, d\}^{ct}$.
We compare CBP and TCBP with regards to number of parameters and computational efficiency in Table~\ref{table:cbp_tcbp}.

We wish to assert that the use of CBP as well as TCBP for multimodal feature encoding is novel.
We learn a projection function to model multimodal data that includes text, images, audio and video.





\begin{algorithm}[t]
\caption{Temporal Compact Bilinear Pooling}\label{algo:ttsp}
\begin{algorithmic}
\STATE \textbf{Input:}  Aggregated feature maps of multimodal data $x \in \mathbb{R}^{c \times t}$,
where $c$ and $t$ are the number of channels and temporal segments.
\STATE \textbf{Output:} Multimodal encoded feature map $v \in \mathbb{R}^{d}$, where $d$ is the encoded feature dimension.
\STATE \textbf{Procedure:}
\STATE \textbf{1.} Initialize random vector $h_{k} \in \{1, \ldots, d\}^{c}$, and $s_{k} \in \{-1, 1\}^{c \times t}$  from a uniform distribution, $k=1,2$.
\vspace{1mm}
\STATE \textbf{2.} Compute the count sketch projection function $u = \Psi(x,h,s) = \{(Px)_{1},\ldots(Px)_{d}\}$, where $(Px)_{j} = \sum_{i:h[i]=j} \sum_t s[i,t] \cdot x[i, t]$.
\vspace{1mm}
\STATE \textbf{3.} Finally, compute the output encoded vector as $v = FFT^{-1}(FFT(\Psi(x,h_{1},s_{1})) \circ FFT(\Psi(x,h_{2},s_{2})))$, where $\circ$ denotes element-wise multiplication operator.
\end{algorithmic}
\end{algorithm}

\begin{table}[t] 
\begin{center}
\resizebox{8cm}{!} {
\begin{tabular}{lcc}
\toprule
	&CBP	& TCBP \\
\midrule
Input dimensions    & $\mathbb{R}^{c}$ & $\mathbb{R}^{c \times t}$ \\
Output dimensions	    &$d$ & 	$d$\\
Parameters ($h, s$)	    &$2 \cdot 2c$ & $2 \cdot (c + ct)$\\
Computation     	    &$O(c + d \log d)$ & $O(ct + d \log d)$ \\
\bottomrule
\end{tabular}}
\end{center}
\vspace{-0.5cm}
\caption{\small{Dimension, memory, and computation comparison of bilinear pooling with tensor sketch~(CBP), and the proposed temporal compact bilinear pooling~(TCBP) algorithm. Parameters $c,t,d$ denote the number of channels, temporal segments, and the projected dimension.}}
\label{table:cbp_tcbp}
\vspace{-0.3cm}
\end{table}





\subsection{Learning via Temporal Ordering}
We wish to learn a multimodal embedding for video clips using temporal ordering as a proxy task.
For example, consider a scene from movie that consists of $M$ video clips, $(V_1, \ldots, V_M)$.
These clips are ordered by the video creators to tell a story in the most natural way, and often encode general principles of causality (e.g., pushing a car gets it rolling) or temporal progression (e.g., enter the house before removing coat).

\vspace{1mm}
\noindent\textbf{Training.}
We sample an ordered pair of clips $(V_i, V_j)$ randomly from the entire scene and train our model parameters to learn that $V_j$ should apear \emph{after} $V_i$, denoted as $V_i \succ V_j$.
To ascertain this order rule, we are inspired by the order violation loss~\cite{vendrov2015order} used to encode visual semantic hierarchy.
In our case, we encode the loss between a pair of clips as:
\begin{equation}
\label{eq:1}
L(V_i, V_j) = \| \max \left(0, \phi_t(V_i) - \phi_t(V_j) \right) \|^2 \, ,
\end{equation}
where $\phi_t(V) \in \mathbb{R}_+$ is the clip representation used for temporal ordering, and the loss expects the feature $\phi_t(V_i)$ to appear ``before'' (below and to the left of in a 2D space) $\phi_t(V_j)$.
All parameters are updated by back-propagating this loss through the network.
Fig.~\ref{fig:pipeline} illustrates this learning procedure for a pair of video clips.

\vspace{1mm}
\noindent\textbf{Inference.}
We are also interested in understanding how well our model is able to order a set of $M$ clips.
To do this, we first encode all $M$ clips using the temporal order feature encoding $\phi_t(V)$, and compute the ordering loss between all $M \cdot (M-1)$ pairs of clips in the scene.
Note that, there are $M!$ possible sequences in which $M$ unordered clips can be sorted.
We use a simple brute-force approach and pick the sequence that results in the smallest overall pairwise ordering loss.
The final ordered list of clips is said to be correct, iff it fully matches the ground-truth ordered list.
While this approach does not scale well, it is a feasible and quick solution when the maximum number of clips in a scene is limited to $M_{\max} = 6$.
We will explore other sorting and ranking techniques such as~\cite{sortingseq} in the future.

\subsection{Implementation details}
\label{sec:details}
We present details of our approach for representing individual clips, their modalities, and other feature extraction procedures below.

\vspace{1mm}
\noindent\textbf{Sampling temporal segments from a clip.}
We observe that most clips in our dataset have 4 to 6 temporal segments (see Table~\ref{table:dataset}), with a maximum of 11.
As our projection function depends on the number of temporal segments (recall that $s$ is randomly initialized to be $\{-1,1\}^{c \times t}$), we choose to use $t=3$ segments to represent all clips.
We consider three sampling strategies to pick a consecutive sequence of three temporal segments ($t=3$) from each clip.
(i) Random Sampling involves randomly choosing three consecutive segments from a clip;
(ii) Constrained First (C\_First) picks the first three segments; and
(iii) Constrained Last (C\_Last) picks the last three segments in the clip.
We observe that C\_Last tends to achieve slightly higher performance and choose this as our default sampling strategy, unless otherwise stated.

\vspace{1mm}
\noindent\textbf{Base representations from pre-trained models.}
Given a video clip, we first break it into non-overlapping 16-frame video segments. 
We extract a variety of feature descriptors for different modalities and discuss their details below:
(1) Object features~(\textbf{I}) in the frame are extracted using a ResNet50 pre-trained on ImageNet dataset~\cite{imagenet};
(2) Place/Scene related features~(\textbf{P}) are obtained using a ResNet50 pre-trained on Places365 dataset~\cite{places};
(3) Video features encoding motion~(\textbf{R}) are obtained using 3D ResNet50~\cite{resnet3d} pre-trained on Kinectics 400 dataset;
(4) Text features~(\textbf{S}) are computed using Glove6B~\cite{glove} pre-trained word vectors;
and finally (5) Audio features~(\textbf{A}) are extracted using the SoundNet~\cite{soundnet} model pre-trained on millions of Flickr videos.

The input to the  3D ResNet50 model is a complete segment of 16 frames with size $112 \times 112 \times 16$.
From this, we extract \texttt{avgpool} features resulting in $\mathbb{R}^{2048}$.
For frame-level features (ImageNet and Places), we first resize the frame to $224\times224$ and extract \texttt{avgpool} features resulting in $\mathbb{R}^{2048}$ per frame.
We then mean pool across 16 frames within the segment to obtain a single feature map of $\mathbb{R}^{2048}$.

To process audio, we first extract the part corresponding to the 16 visual frames (estimated using timestamps) from the entire audio clip.
This is passed to the SoundNet model, and we extract $pool5$ features for the audio-duration corresponding to the 16 frames.
After an averaging across channels,
we obtain a feature map of $\mathbb{R}^{256}$.

The text modality consists of closed captions and subtitles.
Note that we do \emph{not} use manually curated video captions or any other descriptions that may not available for most videos.
We use the 300-dimensional word embeddings from the Glove6B~\cite{glove} model, and mean pool all words that correspond to the clip to obtain a $\mathbb{R}^{300}$ feature map.
As splitting words across temporal segments is non-trivial, we replicate this same feature to all temporal segments.

In summary, a video segment has dimensions~($c$) as follows: \textbf{I}, \textbf{P}, \textbf{R} are 2048, \textbf{A} is 256, and \textbf{S} is 300.
The joint concatenated features are \textbf{PI}: 4096,  \textbf{API}: 4352, \textbf{APIS}: 4652, and \textbf{APIR}: 6400.

\vspace{1mm}
\noindent\textbf{Clip representations.}
For both CBP and TCBP encoding methods, we use the same network architecture.
The input is a feature map $x \in \mathbb{R}^{c\times t}$ consisting of a combination of modalities discussed above.
We first apply a $1\times 1$ convolution on the channels to obtain a $\mathbb{R}^{2048\times t}$ matrix.
This is fed to the encoding method resulting in $v_{\text{enc}} \in \mathbb{R}^{d}$, where $d=8192$.
We then perform feature normalization via a signed square-root operation followed by $\ell_2$ normalization.
We stack a linear layer to obtain a generic clip representation (that can be used for other tasks) as $v_{\text{clip}} = W_1 \cdot v_{\text{enc}} \in \mathbb{R}^{4096}$.
Finally, we compute the temporal ordering feature as $\phi_t(V) = | W_2 (\text{ReLU}( v_{\text{clip}} )) |$, where $W_2 \in \mathbb{R}^{4096 \times 2048}$.
Note that the absolute value $\phi_t(V) \in \mathbb{R}^{2048}_+$ is used to compute the ordering loss.
Biases are present for all linear layers, but are omitted for brevity.

\vspace{1mm}
\noindent\textbf{Training and optimization.}
Our model is implemented in PyTorch v0.4.
We use mini-batch SGD to learn the model parameters with a fixed weight decay of $5 \times 10^{-4}$, momentum of 0.9, a learning rate of $10^{-3}$ and a batch size of 32 for training our network.
Depending on the variant, we train our model for 5000-8000 iterations.
For efficiency reasons, we do not finetune networks used to extract features.

\section{Evaluation}
\label{sec:exp}

In this section, we first introduce the dataset used to train temporal ordering models.
We then show the applicability of the proposed methods on three challenging tasks:
1) inferring the temporal order for a set of unordered video-clips;
and 2) using TCBP for action recognition task.

\subsection{Dataset}
To the best of our knowledge, temporal ordering of videos is a new task.
Thus, we build a new dataset to train and evaluate our approach.
We choose to use clips from the \textit{Large Scale Movie Description Challenge}~(LSMDC)~\cite{lsmdc1,lsmdc2,lsmdc3} dataset as they are publicly available, contain several modalities, and are sourced from movies that typically care about temporal order.
The LSMDC dataset consists of 202 movies with 118,081 video clips.
These clips are associated with transcribed captions from the Audio Descriptions that help visually impaired people understand the visual content on screen, however, we do not use these captions as our text modality as they may not be available for all videos.
Having clips corresponding to audio descriptions also implies that most clips contain none to little dialog.
Additionally, the LSMDC clips are not contiguous, and there are often small (few second) and large (few minutes) gaps between clips.
We extend this dataset by first grouping clips into scenes.

\vspace{1mm}
\noindent\textbf{Scene boundary detection.}
In this task, our goal is to assign clips that are temporally continuous, happen in the same place, or contain a group of coherent events, to the same scene.
We will learn to order a set of clips within a scene.

Scene boundary detection for full movies is usually performed after shot boundary detection.
However, as the LSMDC dataset already comes with pre-segmented clips, we redefine the problem as grouping clips into unique scenes.
We first use a dynamic programming algorithm based on~\cite{tapaswi2014dpscenes} to efficiently assign clips to scenes.
However, scenes that are created using this method are often quite long and contain between 15-30 clips.
In order to segment such scenes into smaller sub-sequences, we use the FINCH clustering algorithm~\cite{finch} that automatically determines the number of clusters.
We manually verify these clip clusters for consistency, and only scenes that contain between 2-6 clips are retained.

We use the same training, validation, and test splits, as used in the LSMDC set of captioning and retrieval tasks.
In total, there are 25,269 scenes in the training set, 1,784 scenes in the validation set, and 2,443 scenes in the test set.
The dataset statistics are given in Table~\ref{table:dataset}.
The top half of the Table presents the total number of clips that appear in scenes of size 2, 3, etc.
In the bottom half we indicate the number of temporal segments (or number of frames) for different videos, also in the range of 2-6 segments.
We divide each clip into $t$ 16-frame segments without overlap, \emph{i.e.}~a clip with $F$ frames has $t = \lceil F/16 \rceil$ segments.
Approximately 65\% of scenes have 4 segments, and there are no clips with less than 48 frames (i.e. with t=3).
At a video framerate of 24 fps, this puts most clips in the range of 2-4 seconds.

We will make our annotations publicly available for further research into the problem of temporal ordering.

\begin{table}[t] 
\begin{center}
\resizebox{8cm}{!}{
\begin{tabular}{lcccccc}
\toprule
\multicolumn{7}{c}{\textbf{\#Clips in a scene}}		\\	
Scene size     &2	&3	&4	&5	&6	& 2-6 \\
\midrule
Training		&13455	&6711	&3097	&1382	&624   &\textbf{25269} \\
Validation		&958	&472	&203	&100	&51	   &\textbf{1784}\\
Test	    	&1333	&588	&325	&135	&62	   &\textbf{2443}\\
\midrule
\midrule
\multicolumn{7}{c}{\textbf{\#Temporal segments}} \\
t	&2	&3	&4	&5	&6	 & 2-6\\
Frames	&32	&48	&64	&80	&96	 & 32-96\\
\midrule
Training		&0	&0	&17992	&3102	&2025	 &\textbf{25269}\\
Validation		&0	&0	&1257	&233	&143	 &\textbf{1784}\\
Test	    	&0	&0	&1542	&382	&252	 &\textbf{2443}\\
\bottomrule
\end{tabular}}
\end{center}
\vspace{-0.5cm}
\caption{Number of clips in a scene of variable size and number of temporal segments in each such clip for our dataset.}
\label{table:dataset}
\vspace{-0.3cm}
\end{table}






\begin{table}[t] 
\begin{center}
\tabcolsep=0.12cm
\resizebox{8cm}{!} {
\begin{tabular}{ll cccc}
\toprule
&\multicolumn{5}{c}{\textbf{Scenes with 2-6 clips and t = 3}}		\\	
Split	& Modality & C\_First& C\_Last & Random Sampling & Chance \\
\midrule
\multirow{5}{*}{Validation}
&A      &22.90 &57.97 &57.21  &31.78 \\
&P      &26.40 &76.35 &73.93 &31.78 \\
&I      &22.42 &77.32 &74.57 & 31.78\\
&R      &39.13 &28.08 &26.90 &31.78 \\
&S      &23.79 &22.74 &21.70 & 31.78 \\ 
\midrule
\multirow{4}{*}{Validation}
&PI     &30.16 &\textbf{79.65}    &71.86  &31.78\\
&API    &24.56 &\textbf{78.70}    &73.15  &31.78\\
&APIS   &21.47  &77.39 &71.33  & 31.78\\
&APIR   &17.04  &77.24	&46.92	&31.78\\
\midrule
\multirow{4}{*}{Test}
&PI	    &34.10     &{81.13} &79.39 &31.89 \\
&API	&39.13    &\textbf{85.06} &\textbf{84.57} &31.89\\
&APIS   &38.41      &83.24          &82.11               &31.89 \\
&APIR   &25.01      &{82.48} &79.38 &31.89 \\
\bottomrule
\end{tabular}
}
\end{center}
\vspace{-0.5cm}
\caption{Feature importance for CBP with different sampling strategies. A, P, I, R, and S denote the Audio, Places, ImageNet, Video and Text modality respectively.}
\label{table:modality_study}
\vspace{-0.5cm}
\end{table}

\subsection{Temporal Ordering} 
In this section, we present results on the temporal ordering task.
We first present an ablation study for design choices of our proposed method, and then compare our method against other baselines.
We use the stringent ordering accuracy metric that counts a sample as correct only when all video clips are  sorted in the correct sequence.

\vspace{1mm}
\noindent\textbf{Study of individual and combined modalities.}
In this evaluation, we explore the contribution of individual and combined feature representations for temporal ordering. 
For the individual modalities, we do not perform channel reduction with the $1 \times 1$ convolution, as the dimensions are often already small (e.g., $c = 256$ for audio).
In Table~\ref{table:modality_study}, we report the performance of using individual or joint modalities.
Audio~(A), Places~(P), and ImageNet~(I) modalities seem important for this task, while the joint PI and API feature representations perform best.
It is interesting to see that the video features~(APIR) and  word embeddings~(APIS) reduce performance as compared to API.
We believe that motion features captured by R are not very strong to predict what may happen far into the future, while the sparsity of subtitles hurts the text modality.

\vspace{1mm}
\noindent\textbf{Study on ordering a variable number of clips in a scene.}
In our dataset, we have scenes with variable number of clips ranging from 2-6. 
In Table~\ref{table:S_study}, we show the performance of different modalities on subsets of scenes that contain a fixed number of clips.
We observe similar performance from PI and API for scenes with 2-4 clips.
However, API is able to correctly order scenes with 5 and 6 clips, while PI fails.
The failure cases show that in absence of varying objects, context and motion, the model fails to predict the future due to intrinsic ambiguity of stationary scenes, while audio may provide some help.

We choose API as the default modality for all further experiments.

\begin{table}[t] 
\begin{center}
\tabcolsep=0.15cm
\resizebox{8cm}{!} {
\begin{tabular}{ l ccccc c }
\toprule
\# Clips in scene	&2	&3	&4	&5	&6 &2-6\\
Samples &958	&472	&203	&100	&51	   &1784\\
\midrule
Random       & 50.00 & 16.67 & 4.17 & 0.83 & 0.14 & 31.78 \\
PI		     &96.86 &60.38 &91.13 &22.00 &1.96 & \textbf{79.65}\\
API		 &95.51 &56.14 &87.19 &31.00 &31.37 & \textbf{78.70}\\
APIS	    &98.28	&54.01	&76.34	&16.45	&25.13	&77.39 \\
APIR	     &100 &42.58 &88.18 &40.00 &0.0 &77.24\\
\bottomrule
\end{tabular}}
\end{center}
\vspace{-0.5cm}
\caption{Impact of the number of clips per scene on ordering accuracy on the validation split. Encoding method used is CBP with C\_Last sampling strategy.}
\label{table:S_study}
\vspace{-0.3cm}
\end{table}

\begin{table}[t] 
\begin{center}
\resizebox{6cm}{!} {
\begin{tabular}{ l ccc }
\toprule
Number of segments	&$t = 1$	&$t = 2$	&$t = 3$	\\
\midrule
PI		&71.14	&77.30	    &\textbf{79.65}	   \\
API		&54.88 &\textbf{77.92}     &\textbf{78.70} \\   
APIS	&53.48	&76.19	&77.39	\\
APIR	&71.08 &77.19     &77.24   \\
\bottomrule
\end{tabular}}
\end{center}
\vspace{-0.5cm}
\caption{Impact of the number of temporal segments used in the encoding of video clips on ordering accuracy on the validation split.
Encoding with CBP with C\_Last sampling.}
\label{table:L_study}
\vspace{-0.5cm}
\end{table}

\vspace{1mm}
\noindent\textbf{Study of variable temporal segments encoding.}
In this experiment, we vary the number of temporal segments that are encoded for each clip. Recall each segment $t$ represents a chunk of 16 frames, and clips in our dataset range from 4 to 11 segments.
We use $t = 3$ for most experiments as it acts as a kind of data augmentation when using random sampling.Table~\ref{table:L_study} confirms our expectation that using more segments leads to better clip representations improving performance for all modalities.

\vspace{1mm}
\noindent\textbf{Comparison of CBP vs. TCBP.} Table~\ref{table:cbp_vs_tcbp} shows the performance comparison for variable number of temporal segments.
Note that when $t = 1$, CBP and TCBP are identical.
However, for $t = 2, 3$, TCBP shows slight improvements over CBP as it is better able to encode longer clips, while CBP relies on sum pooling.

\begin{table}[t] 
\begin{center}
\resizebox{6cm}{!} {
\begin{tabular}{ ll ccc }
\toprule
Split & Method & $t = 1$ & $t = 2$ & $t = 3$ \\
\midrule
\multirow{2}{*}{Validation}
&  CBP & 54.87 & 77.92 & 78.70 \\
& TCBP & 54.87 & 78.13 & \textbf{79.43} \\
\midrule
\multirow{2}{*}{Test}
&  CBP & 68.65 & 84.69 & 85.06 \\
& TCBP & 68.65 & 85.26 & \textbf{86.23} \\
\bottomrule
\end{tabular}}
\end{center}
\vspace{-0.5cm}
\caption{Comparison between CBP and TCBP with C\_Last sampling. Modalities API are used in this experiment.}
\label{table:cbp_vs_tcbp}
\vspace{-0.3cm}
\end{table}

\begin{table}[t] 
\begin{center}
\resizebox{5cm}{!} {
\begin{tabular}{ l cc }
\toprule
API &	Validation &Test 	\\
\midrule
Chance          &31.78      &31.89 \\
ConcatT\_MLP    &19.78      &37.17 \\
Mean\_Pool      &66.59     &69.58 \\
NetVLAD         &68.11      &83.79 \\
\midrule
CBP             &78.70     &85.06 \\
TCBP            &\textbf{79.43} &\textbf{86.23} \\
\bottomrule
\end{tabular}}
\end{center}
\vspace{-0.5cm}
\caption{Comparison of TCBP with other commonly used encoding strategies using C\_Last sampling. Modalities API are used here.}
\label{table:soa}
\vspace{-0.3cm}
\end{table}

\begin{table}[t] 
\begin{center}
\resizebox{5cm}{!} {
\begin{tabular}{ l cc }
\toprule
API & Validation & Test  	 	\\
\midrule
NetVLAD & 68.11 & 83.79\\
NetVLAD w Neg. & 71.21 & 84.99 \\
\midrule
CBP         &78.70 &85.06\\
CBP w Neg.  &84.59 & 88.95 \\
\midrule
TCBP        &79.43 & 86.23\\
TCBP w Neg. &\textbf{83.18} & \textbf{89.19}\\
\bottomrule
\end{tabular}}
\end{center}
\vspace{-0.5cm}
\caption{Role of negative mining in CBP and TCBP using C\_Last sampling. Modalities API are used here.}
\label{table:negatives}
\vspace{-0.5cm}
\end{table}

\vspace{1mm}
\noindent\textbf{Comparison between TCBP and other encoding methods.}
We compare TCBP with several popular pooling methods.
As before, we start with an input feature map $x \in \mathbb{R}^{c \times t}$.
\textbf{ConcatT\_MLP:} first applies $1 \times 1$ convolutions with 2048 kernels, resulting in $\mathbb{R}^{2048 \times t}$.
We then simply flatten the temporal segments and concatenate them to form a vector of size $\mathbb{R}^{2048 \cdot t}$.
\textbf{Mean\_Pool:} We average pool across the temporal segments $t$, resulting in a vector of $\mathbb{R}^{c}$.
\textbf{NetVLAD:} we first apply $1 \times 1$ convolutions with 512 kernels, resulting in a feature map of $\mathbb{R}^{512 \times t}$.
This is fed to NetVLAD~\cite{vladnet1,actionvlad,miech2017learnable} with 32 clusters\footnote{We optimized the hyperparameters of NetVLAD by evaluating \#cluster=\{16,32,64\}, feat\_dim=\{256,512,1024\}, and found that for \#cluster=32 and feat\_dim=512 resulted in the best performance.} resulting in a feature vector of $\mathbb{R}^{16384}$. 

All these methods (except Mean\_Pool) are followed by two fully-connected layers yielding hidden states of $\mathbb{R}^{4096}$ and $ \mathbb{R}^{2048}$.
For Mean\_Pool we use $\mathbb{R}^{2048}$ and $\mathbb{R}^{1024}$ as the dimensions are already small.
The output of the last linear layer is used in the ordering loss.

In Table~\ref{table:soa}, we see that TCBP outperforms all methods, including strong baselines of CBP and NetVLAD.
In fact, surprisingly ConcatT\_MLP performs very poorly, while a simple Mean\_Pool shows limited performance.
Note that we expect small improvements for TCBP over CBP as the main difference lies in the projection of temporal segments directly instead of sum pooling over time after projection as in CBP.
However, we believe that TCBP is a principled way to handle temporal segments within the bilinear pooling framework.

\vspace{1mm}
\noindent\textbf{Role of negative mining.}
So far, we have only considered training our model with positive examples~($Pos$) of ordered video clips.
Now, we analyze the impact of also including negative examples~($Neg$).
We adopt a popular negative sampling strategy~\cite{socher2013reasoning}, and corrupt each positive pair by replacing one of the clips with a randomly chosen clip from the same batch.
The overall loss is:
\begin{equation}
\label{eq:2}
\Lb = \sum_{(V_i,V_j) \in Pos} L(V_i, V_j) + \sum_{(V_i,V_j') \in Neg} \max(0, \alpha - L(V_i, V_j'))  \, ,
\end{equation}
where $L(\cdot, \cdot)$ is the ordering loss (see Eq.~\ref{eq:1}), $\alpha$ is a margin (set to 0.2), and $(V_i, V_j')$ is the corrupted version of $(V_i, V_j)$ without loss of generality by only replacing $V_j$.
We use an equal ratio of positive to negative examples to train our model.

In Table~\ref{table:negatives}, we observe that adding negatives improves the performance for all methods as compared to training the model with positive examples only.
Note that we do not need to necessarily rely on negative examples, and positive examples are sufficient to learn a good multimodal representation as seen in Table~\ref{table:soa}. Nevertheless, it is good to see that further improvements can be achieved by including negatives.


\subsection{Action Recognition}
In our last experiment, we evaluate the effectiveness of TCBP for video action recognition task. 

\vspace{1mm}
\noindent\textbf{Dataset.}
We evaluate TCBP on two human action and activities datasets; HMDB51~\cite{hmdb51} and UCF101~\cite{ucf101}.
The HMDB51 dataset consists of 51 action categories with 6,766 video clips.
The UCF101 dataset consists of 101 action classes with 13,320 video clips.
We report mean average accuracy over the three pre-defined splits for both datasets.

\vspace{1mm}
\noindent\textbf{Method.}
We use TLE~\cite{tle} as the base architecture, and employ TCBP as the encoding method.
In particular, we  use the  C3D ConvNet~\cite{c3d} pre-trained on the Sport-1M dataset~\cite{karpathy}.
The network operates on a stack of 16 RGB frames of size $112\times112$. Following the setup of Diba~\etal~\cite{tle}, we employ TSN~\cite{tsn} architecture with 3 segments pushed through the C3D ConvNet.
Same as~\cite{tle}, we extract the convolutional feature maps from the last convolutional layers and feed them as input to TCBP.
This is followed by a classification layer with a $K$-way softmax,  where $K$ is  the  number  of action  categories.
Note that convolutional feature maps from each segment of the last layer of C3D produce an output of size $\mathbb{R}^{512\times2\times7\times7}$.
Three such segments are first aggregated via element-wise multiplication (similar to TLE), and followed by TCBP projection with $t=2$ resulting in a vector of $\mathbb{R}^{8192}$ dimensions.
We use the same training and evaluation setup as \cite{tle} for a fair comparison.

\vspace{1mm}
\noindent\textbf{Results.}
Table~\ref{table:action} shows the performance of TCBP with C3D ConvNets compared against several other methods.
The goal of this experiment is to show that TCBP can improve the performance of the base C3D~\cite{c3d}.
It is also interesting to see that, TCBP improves the C3D ConvNet performance over the two-stream ConvNets~\cite{twostream} on both datasets.
Further, TCBP outperforms Bilinear pooling, CBP~\cite{tle}, and iDT+FV~\cite{idt} by significant margin.
We believe that the reason for TCBP outperforming other methods is its ability to effectively encode temporal cues in the video data, which are not considered in other methods such as~\cite{tle,twostream,c3d,idt}. 

Furthermore, in practice, NetVLAD has been shown to perform worse than Bilinear pooling~(e.g., CBP)  in~\cite{tle,actionvlad}.
This is most likely due to CBP capturing interactions between features at each channel, thus leading to a strong representation.


\begin{table}[t] 
\begin{center}
\resizebox{8cm}{!}{
\begin{tabular}{ l c c }
\toprule
Method &  UCF101 & HMDB51\\
\midrule
SpatioTemporal ConvNet~\cite{karpathy}	& 65.4 & $-$	\\ 
LRCN~\cite{lrcn}					& 82.9 & $-$	\\ 
Composite LSTM Model~\cite{compositelstm}	& 84.3 & 44.0	\\  
iDT+FV~\cite{idt}			& 85.9 & 57.2	\\ 
Two Stream~\cite{twostream}		& 88.0 & {59.4}	\\ 
C3D~\cite{c3d}					& 82.3 & 56.8	\\ 
TLE: Bilinear+TS~(CBP)~\cite{tle}	&	85.6 	&  59.7 \\
TLE: Bilinear~\cite{tle}			& 86.3		&  60.3	\\ 
\midrule
TLE:TCBP~(ours) 				& \textbf{88.03}		& \textbf{61.58} \\
\bottomrule
\end{tabular}}
\end{center} 
\vspace{-0.5cm}
\caption{\textbf{C3D ConvNets.} Comparison of accuracy~(\%) of TCBP with C3D ConvNet against state-of-the-art methods over all three splits of UCF101 and HMDB51.}
\vspace{-0.3cm}
\label{table:action}
\end{table}

\section{Conclusion}
\label{sec:conc}
We proposed a unified network architecture to learn an effective and compact multimodal video clip representation that jointly encodes images, audio, video, and text.
In particular, we proposed TCBP, an extension to Compact Bilinear Pooling that effectively encodes multiple temporal segments of a video clip.
Our model is trained in a self-supervised learning approach where ordering a set of video clips provides necessary supervision.
We evaluated our feature representation for temporal ordering of video clips, 
where, multiple modalities proved to be helpful.
We also demonstrated that TCBP is a strong encoding method that performs well on other video tasks such as action recognition.
We strongly believe that complete understanding of video clips can only be achieved by analyzing all modalities jointly, and hope that such multimodal representations will inspire the community in the future.


\vspace{1mm}
\noindent
\textbf{Acknowledgments.} This work is supported by the DFG (German Research Foundation) funded PLUMCOT project.

{\small
\bibliographystyle{ieee}
\bibliography{main}
}

\end{document}